\documentclass[conference]{IEEEtran}
\IEEEoverridecommandlockouts

\usepackage{cite}
\usepackage{amsmath,amssymb,amsfonts}
\usepackage{algorithmic}
\usepackage{graphicx}
\usepackage{textcomp}
\usepackage{xcolor}
\usepackage{subfigure}
\usepackage{balance}
\usepackage{comment}

\def\BibTeX{{\rm B\kern-.05em{\sc i\kern-.025em b}\kern-.08em
    T\kern-.1667em\lower.7ex\hbox{E}\kern-.125emX}}
\begin{document}

\title{FiDTouch: A 3D Wearable Haptic Display for the Finger Pad\\
}
\author{\IEEEauthorblockN{Daria Trinitatova}
\IEEEauthorblockA{\textit{Intelligent Space Robotics Laboratory} \\
\textit{Skolkovo Institute of Science and Technology (Skoltech) }\\
Moscow, Russia \\
daria.trinitatova@skoltech.ru}
\and
\IEEEauthorblockN{Dzmitry Tsetserukou}
\IEEEauthorblockA{\textit{Intelligent Space Robotics Laboratory} \\
\textit{Skolkovo Institute of Science and Technology (Skoltech)}\\
Moscow, Russia \\
d.tsetserukou@skoltech.ru}

}

\maketitle

\begin{abstract}
The applications of fingertip haptic devices have spread to various fields from revolutionizing virtual reality and medical training simulations to facilitating remote robotic operations, proposing great potential for enhancing user experiences, improving training outcomes, and new forms of interaction. In this work, we present FiDTouch, a 3D wearable haptic device that delivers cutaneous stimuli to the finger pad, such as contact, pressure, encounter, skin stretch, and vibrotactile feedback. The application of a tiny inverted Delta robot in the mechanism design allows providing accurate contact and fast changing dynamic stimuli to the finger pad surface. The performance of the developed display was evaluated in a two-stage user study of the perception of static spatial contact stimuli and skin stretch stimuli generated on the finger pad. The proposed display, by providing users with precise touch and force stimuli, can enhance user immersion and efficiency in the fields of human-computer and human-robot interactions. 
\end{abstract}


\section{Introduction}

Fingertip haptic devices (FHD) enrich the user experience in the realm of human-computer and human-robot interaction, bridging the gap between the digital and physical worlds by providing various cutaneous and force feedback directly to the user's fingertips. The ability to accurately reproduce the feeling of grasping in a virtual or remote environment is essential for creating a realistic experience in Virtual Reality, teleoperation, and telexistence, since finger pads are used for interactions with physical objects and probing the environment in most cases. 

Haptic feedback plays an important role in motion guidance, assisting users to follow a predefined path or trajectory accurately. For example, Gil et al. \cite{gil2022handguidance} explored the approach of providing directional movement cues for guidance in 3D space using a handheld haptic device that delivers shear forces to the finger pads.
The application of haptic feedback during teleoperation provides many benefits in improving task performance and situational awareness of operators. In addition, with the presence of haptic guidance, operators can better
detect obstacles, avoid collisions, and handle fragile or deformable objects with greater precision. Thus, Zhu et al. \cite{zhu2022cutaneous} proposed combining a fingertip interface that provides cutaneous feedback with a desktop force feedback haptic device to increase the dexterity of telemanipulation. In \cite{tsykunov2019swarmtouch}, an approach was introduced for safe guiding of a drone formation using a glove with vibrotactile feedback. Vibrotactile patterns generated at the fingertips are used to provide the operator with information about the state of the swarm, facilitating the control of complex drone formations. By simulating the sense of touch on the fingers during teleoperation, haptic feedback can help operators better understand the remote environment, including the texture, shape, and weight of the objects being manipulated. 

\begin{figure} [t]
\begin{center}
\includegraphics[width=0.95\linewidth]{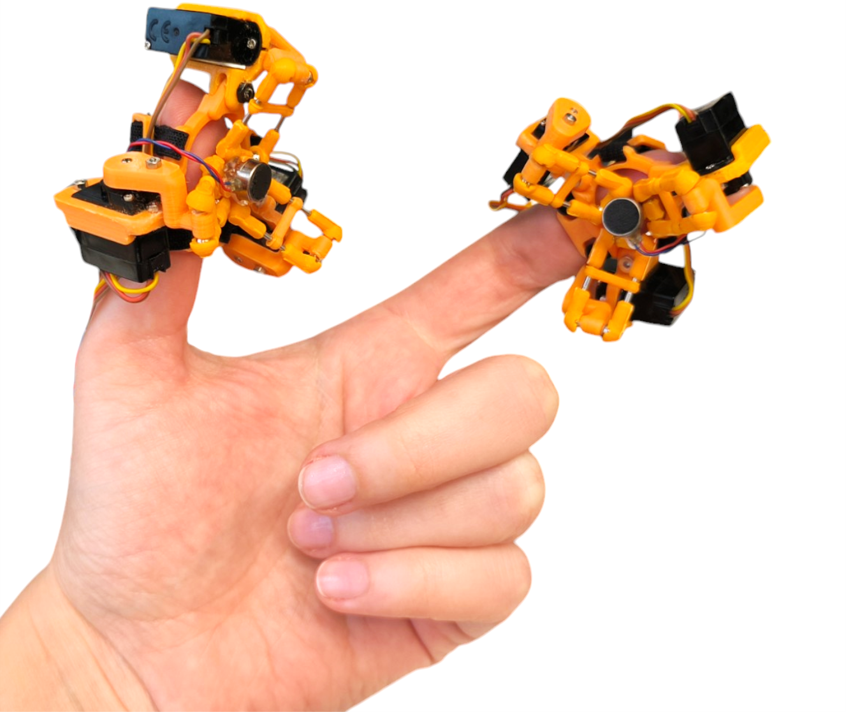}
\caption{A prototype of the 3-DoF haptic display for the finger pads.}
\label{fig:fDeltaTouch}
\end{center}
\end{figure}

In this paper we present the design and evaluation of FiDTouch, a low-cost wearable haptic display with 3-DoF based on the inverted Delta robot (iDelta) that provides cutaneous stimuli to the finger pad such as contact, pressure, encounter, skin stretch, and vibrotactile feedback (Fig. \ref{fig:fDeltaTouch}). We evaluated the performance of the finger-worn tactile display through two user study experiments. In the first user study, we explored the discrimination of contact spatial patterns delivered to the finger pad. In the second experiment, we investigate whether the participant could discriminate between different directions of skin stretch generated by FiDTouch.
 
\section{Related Work }

\subsection{Mechanical stimulation devices}
The most general approach to generating haptic feedback when grasping virtual or remote objects is the stimulation of tactile sensations by touching the fingertips with various mechanical devices, in particular, based on parallel structures \cite{leonardis20173,schorr2017three,lee2018wearable,young2019implementation}. Tzemanaki et al. \cite{tzemanaki2018design} presented a FHD for remote palpation applications based on a variable compliance platform that could be linearly displaced toward the finger pad. 
An interesting approach to combining tactile and kinesthetic feedback was presented in \cite{chinello2019modular}, where a modular wearable interface based on a 3-DoF fingertip cutaneous device combined with a 1-DoF kinesthetic finger module was proposed.
The area of tactile feedback generation for most of the proposed devices is limited to the first phalanx of the finger. The possibility of presenting haptic feedback on two phalanxes of the finger was explored in \cite{linkring}. A designed haptic display with 2-DoF at each contact point can deliver tactile stimuli on distal and middle phalanges.

A relatively novel approach to designing FHDs is to implement folding structures based on the principles of origami. Thus, Williams et al. \cite{williams20214} presented a finger-mounted 4-DoF haptic device to deliver normal, shear, and torsional forces to the fingertip. The parallel kinematic mechanism was created using origami principles. In \cite{haptigami}, a 3-DoF FHD with origami structure was developed to provide cutaneous and vibrotactile feedback. The device has a small size and low weight due to its origami structure. Similarly, the origami principle was used for the design of FingerPrint, a soft FHD with 4-DoF \cite{fingerprint}. The FingerPrint device can generate vibration stimuli, shear and rotational forces on the finger pads, and skin pressure.

\subsection{Electro-tactile stimulation devices}

Electro-tactile stimulation allows high-resolution information to be rendered about grasped surfaces (e.g. texture, shape). Hummel et al. \cite{hummel2016lightweight} introduced a haptic device for multifinger stimulation with electro-tactile feedback. Each electro-tactile tactor comprised an array of eight electrodes. The evaluation user study revealed that the use of electro-tactile feedback improved performance in object manipulation and grasping tasks while reducing the workload demands. Yem et al. \cite{yem2018effect} studied the effect of electrical stimulation on the perception of softness/hardness and stickiness of a virtual object using electrode arrays placed at the fingertip and on the back of the hand. Withana et al.\cite{withana2018tacttoo} introduced Tactoo, an electro-tactile interface with a slim form factor integrated in a temporary tattoo. The prototype of FHD presented consists of an array of eight circular electrodes (2 $mm$ in diameter) with a total activation area of 10 × 10 $mm$. Vizcay et al. \cite{vizcay2021electrotactile} proposed the technique to render the electro-tactile feedback based on pulse width and frequency modulation to improve contact information in VR.

\subsection{Hydraulic and pneumatic actuation}
Fingertip haptic devices based on hydraulic and pneumatic actuation are designed to provide tactile feedback to the user by leveraging fluid (liquid or compressed air) pressure. Thus, Miyakami et al. \cite{miyakami2019hapballoon} presented a finger-worn device generating forces based on balloons inflated through air tubes. The generated pressure feedback is combined with vibration and thermal modules. Han et al. proposed a fingertip-worn ring actuated using a hydraulic system that controls liquid flows to provide pressure, temperature and vibration feedback \cite{han2018hydroring}. Talhan et al. \cite{talhan2022multi} designed a pneumatic thimble-shaped haptic display, made of a silicone membrane with an air cavity to deliver pressure and vibration sensations. Ma et al. \cite{ma2024airpush} developed AirPush, a wearable pneumatic haptic device with 2-DoF delivering adjustable force feedback using compressed air.

\subsection{Finger pad-free haptic devices}

With the wide spread of augmented and mixed reality applications, new design principles of FHD have emerged to avoid interference between user's fingers and physical objects.
In \cite{hring}, an approach was presented to locate a tactile actuator on the proximal finger phalanx to avoid occlusions during hand tracking. A proposed wearable 2-DoF haptic device is able to generate pressure and skin stretch stimuli to the proximal phalanx of the finger. Similarly, a 2-DoF FHD was developed for stiffness simulation of tangible objects in VR/AR environment \cite{de2018enhancing}. 
In \cite{preechayasomboon2021haplets}, a light-weight fingernail-mounted haptic display was introduced that provides vibrotactile feedback. Kawazoe et al. \cite{kawazoe2021tactile} designed Tactile Echoes, a finger-worn haptic device that generates vibrotactile and auditory feedback to the user. The presented system captures touch-elicited vibrations using a piezoelectric sensor and transforms them to tactile feedback for the finger through an inertial voice coil actuator and sound through a loudspeaker or headphone. Maeda et al. \cite{maeda2022fingeret} designed Fingeret, a finger pad-free haptic device consisting of two finger-side actuators (planetary gear motors) and finger nail actuator (vibration LRA) to render low-frequency and high-frequency haptic stimuli. 

An alternative approach to present fingertip sensations is to relocate actuator mechanisms to the forearm. Thus, Moriyama et al. \cite{moriyama2022wearable} proposed a wearable haptic device based on five-bar linkage mechanisms attached to the forearm to present the direction and magnitude of the force on the index finger and thumb. A similar approach was studied in the work of \cite{palmer2022haptic}, where wrist-worn 1-DoF tactile devices provided skin deformation were used to display simulated interaction forces at the fingertips. Tanaka et al. \cite{tanaka2024reawristic} introduced an electro-tactile wristband to provide tactile feedback to the thumb and index finger during touch and pinch interactions in MR.
\begin{figure} [!t]
\begin{center}
    \subfigure[Isometric view.]{
    \includegraphics[width=0.48\linewidth]{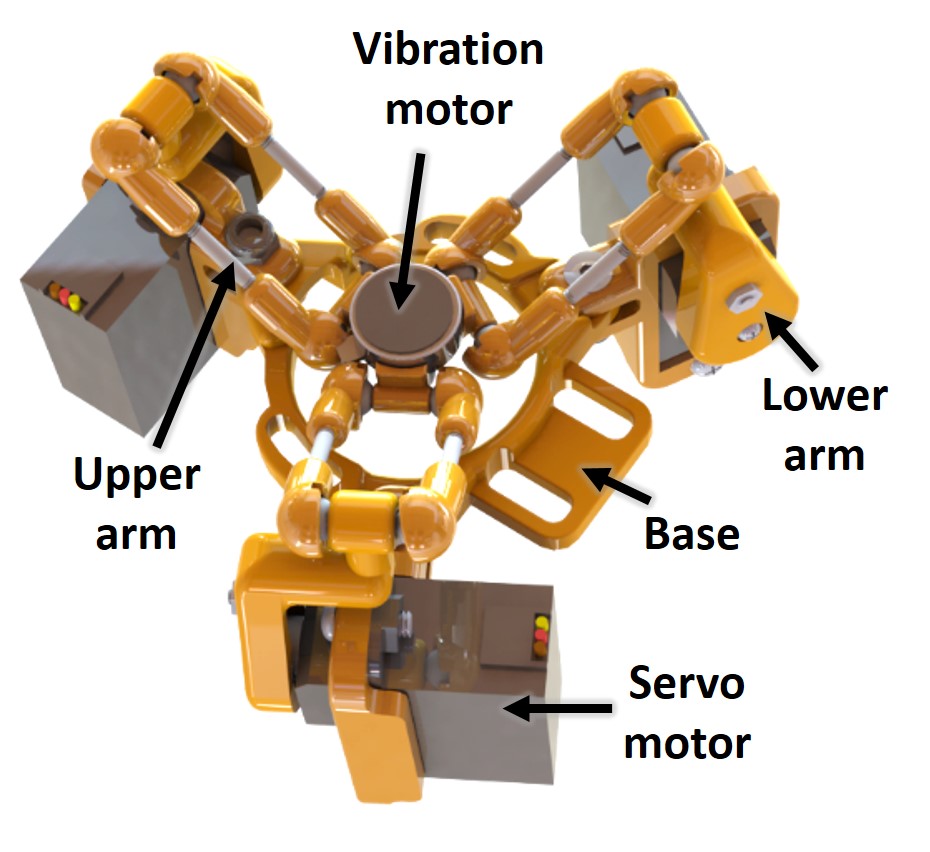}}
    \subfigure[Bottom view.]{
   \includegraphics[width=0.48\linewidth]{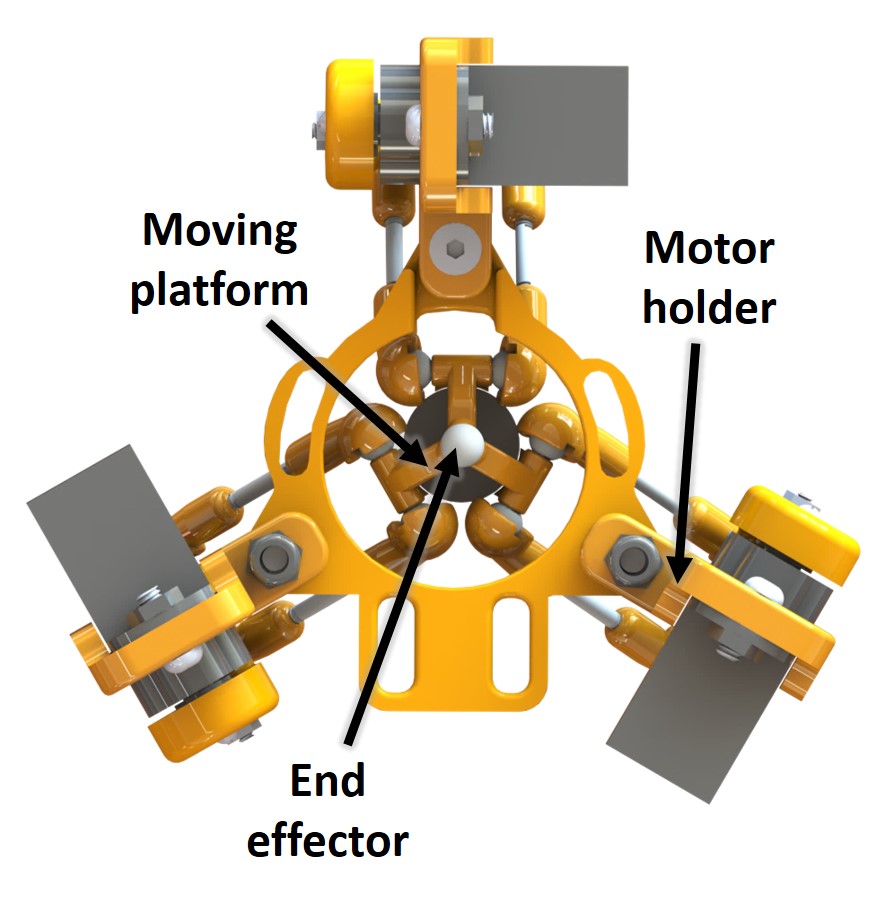}}
\caption{A CAD model of a FiDTouch tactile display.}
\label{fig:fingerDelta_CAD}
\end{center}
\end{figure}

\section{Design of FiDTouch Haptic Display}

The haptic display design is based on the 3-DoF inverted Delta parallel mechanism inspired by the DeltaTouch palm-worn haptic display \cite{deltatouch}, \cite{touchvrr}. The prototype consists of 3D printed parts, namely, motor holders fastened to the base and links made of PLA and stainless steel rods, respectively, and metal spherical joints connecting the links and the moving platform (Fig. \ref{fig:fingerDelta_CAD}). 
The base was designed as an equilateral triangle with a circle hole (\o 16 $mm$ and \o 18 $mm$, for the index finger and thumb, respectively) at the center and two fasteners to the finger pad using elastic tapes. The device is wearing to the distal phalanx in such a way that the end effector translates along almost a flat surface of the finger pad. The end effector is a metal ball with a diameter of 3 $mm$. The moving platform has a base to attach a vibration motor (coin type, \o 8 $mm$) on the top side.

\begin{table}[!h]
\caption{Technical Specifications of the FiDTouch Compared to the Similar FHD with 3-DoF}
\label{spec}
\centering
\includegraphics[width=0.95\linewidth]{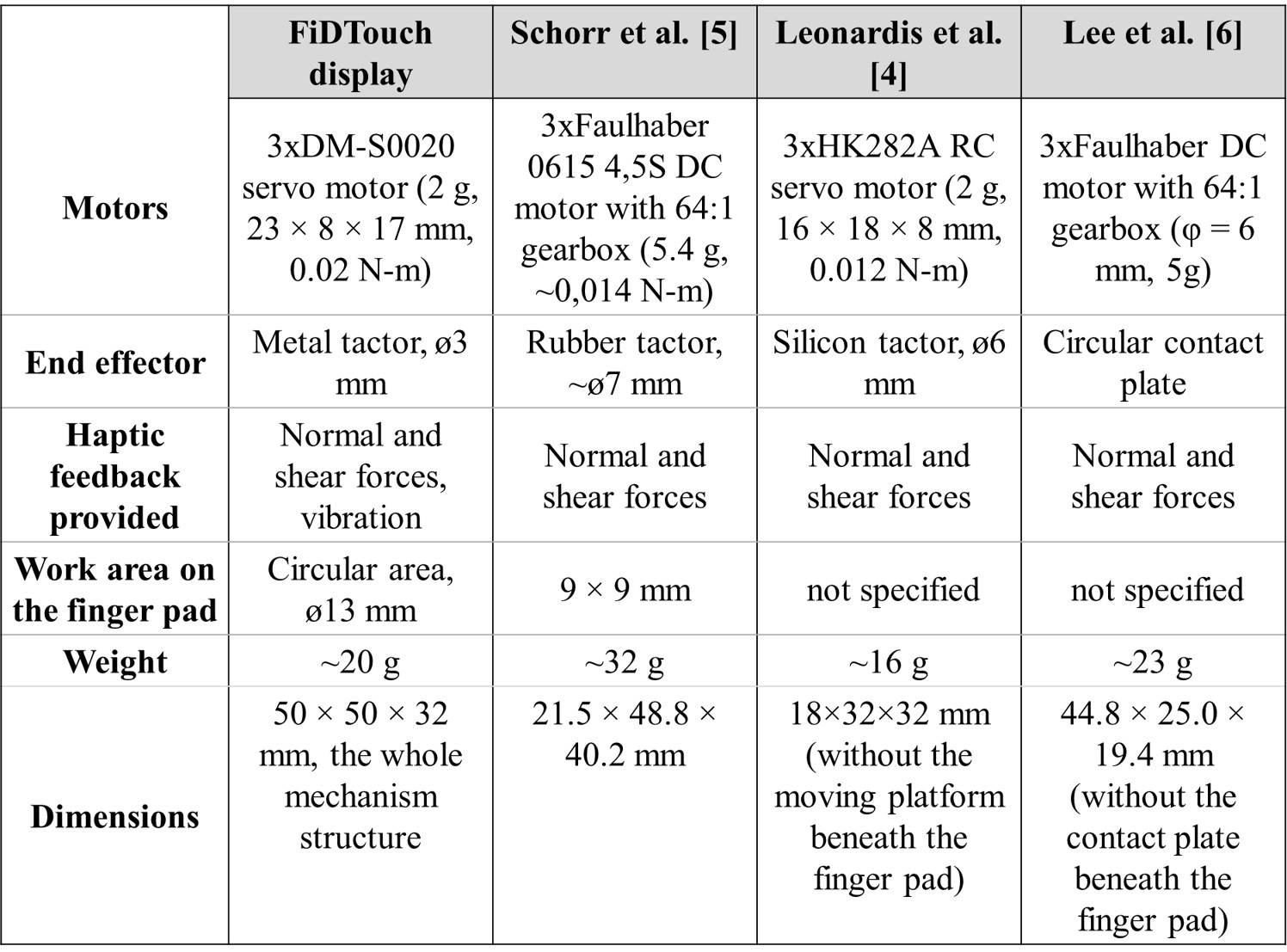}
\end{table}

\begin{figure} [!t]
\begin{center}
    \subfigure[A top view.]{
    \includegraphics[width=0.45\linewidth]{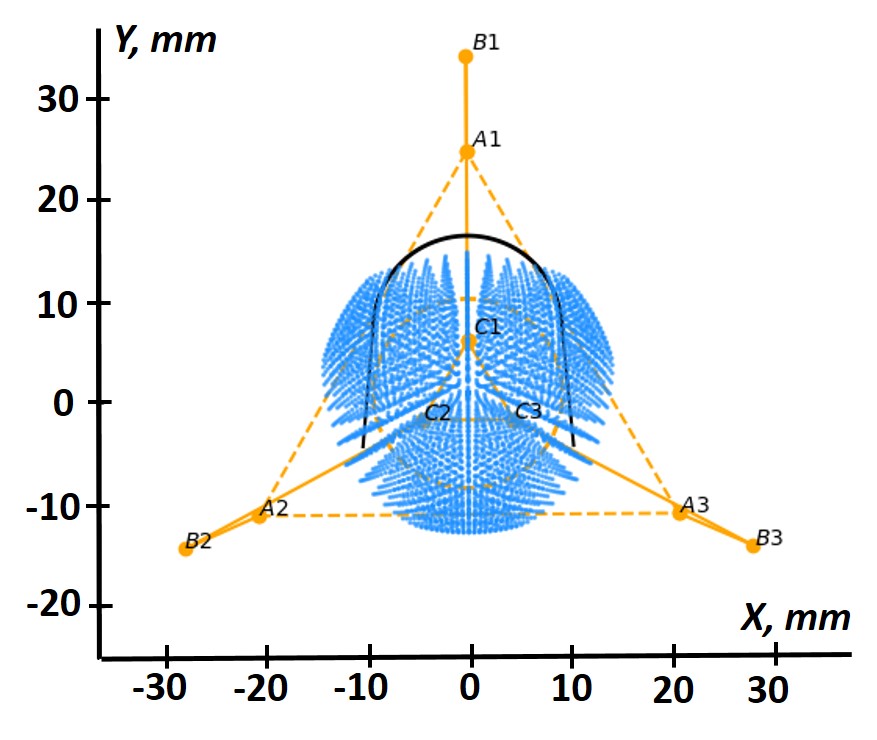}}
    \subfigure[A side view.]{
   \includegraphics[width=0.45\linewidth]{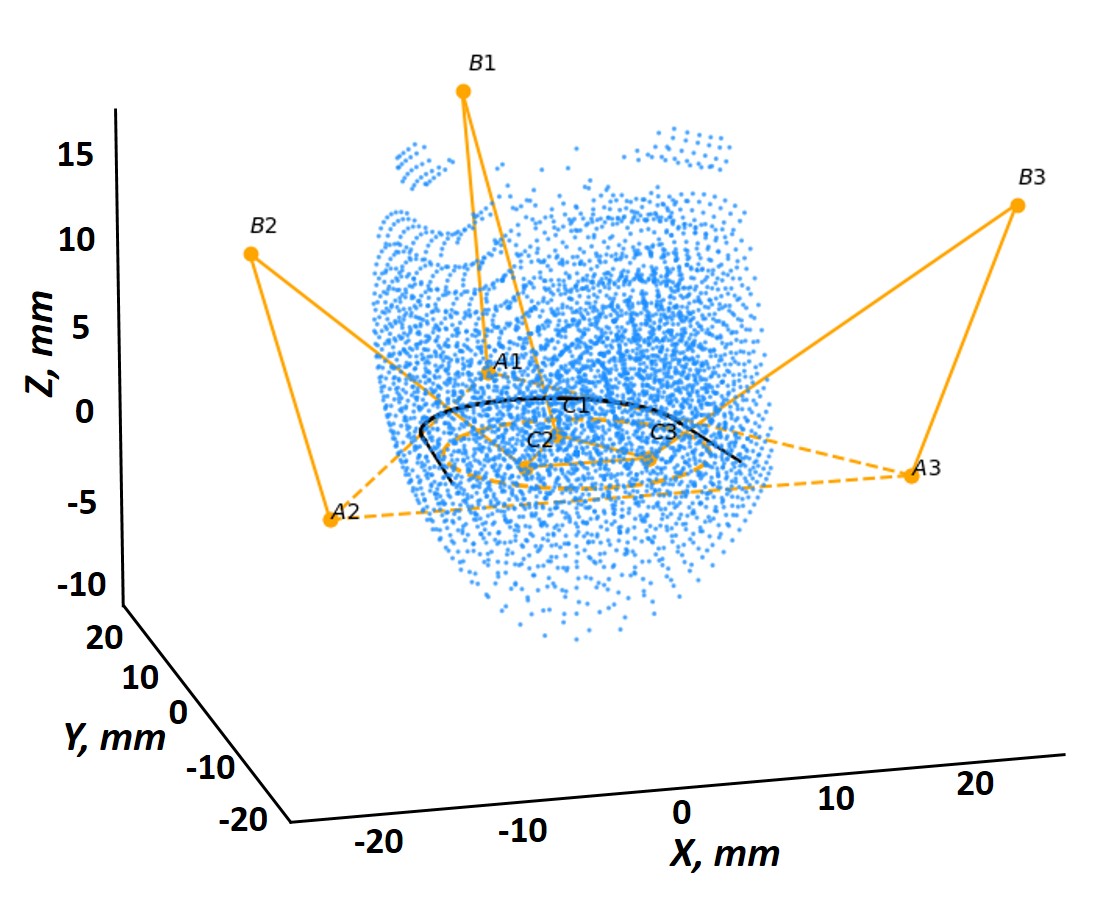}}
\caption{A workspace of FiDTouch display.}
\label{fig:workspace_finger}
\end{center}
\end{figure}

The device is actuated by three DM-S0020 micro servo motors (weight is of 2.1 $g$, dimensions are 23 $\times$ 8 $\times$ 17 $mm$, maximum torque is of 0.2 $kg-cm$) ensuring high force output relative to the lightweight mechanism structure. The servo motors are driven by ESP32 microcontroller and I2C-controlled pulse-width modulation (PWM) driver PCA9685. The specification of the FiDTouch display is shown in Table \ref{spec}. In addition, the characteristics of similar 3-DoF FHDs with parallel mechanism structure are provided. During the development of the FiDTouch, we aimed to provide as large working area on the finger pad as possible while maintaining a compact structure size. Thus, the entire structure of the device (including the moving platform) is enclosed in a frame box of 50 $\times$ 50 $\times$ 32 $mm$. A workspace of the designed inverted Delta mechanism is shown in Fig. \ref{fig:workspace_finger}. Although the implemented inverted Delta parallel mechanism provides a large workspace for end effector movements, we limited the movement of the mechanism to a circular area (\o 13 $mm$) within the base plate with a vertical displacement of  8 $mm$. While all the devices mentioned operate on the same principle, providing contact forces to the finger pad area, the FiDTouch display allows rendering spatially distinct and precise contact pressure with a small ball tactor. Besides, FiDTouch integrates a  vibration motor on the moving platform, enabling combined force and vibrotactile feedback. Compared to the Delta-based FHD presented by Schorr et al. \cite{schorr2017three}, the FiDTouch display provides a larger working area with comparable size and output forces, utilizing servo motors, which are a low-cost and easily controllable alternative to brushless DC motors. 

\subsection{Haptic modes}
FiDTouch display is designed to provide a wide range of tactile stimuli at the finger pad. A summary of the types of tactile stimuli generated by the developed display is presented in Table \ref{tactile_stimuli}.
\begin{table}[!h]
\caption{Types of Haptic Feedback Provided}
\label{tactile_stimuli}
\centering
\includegraphics[width=0.98\linewidth]{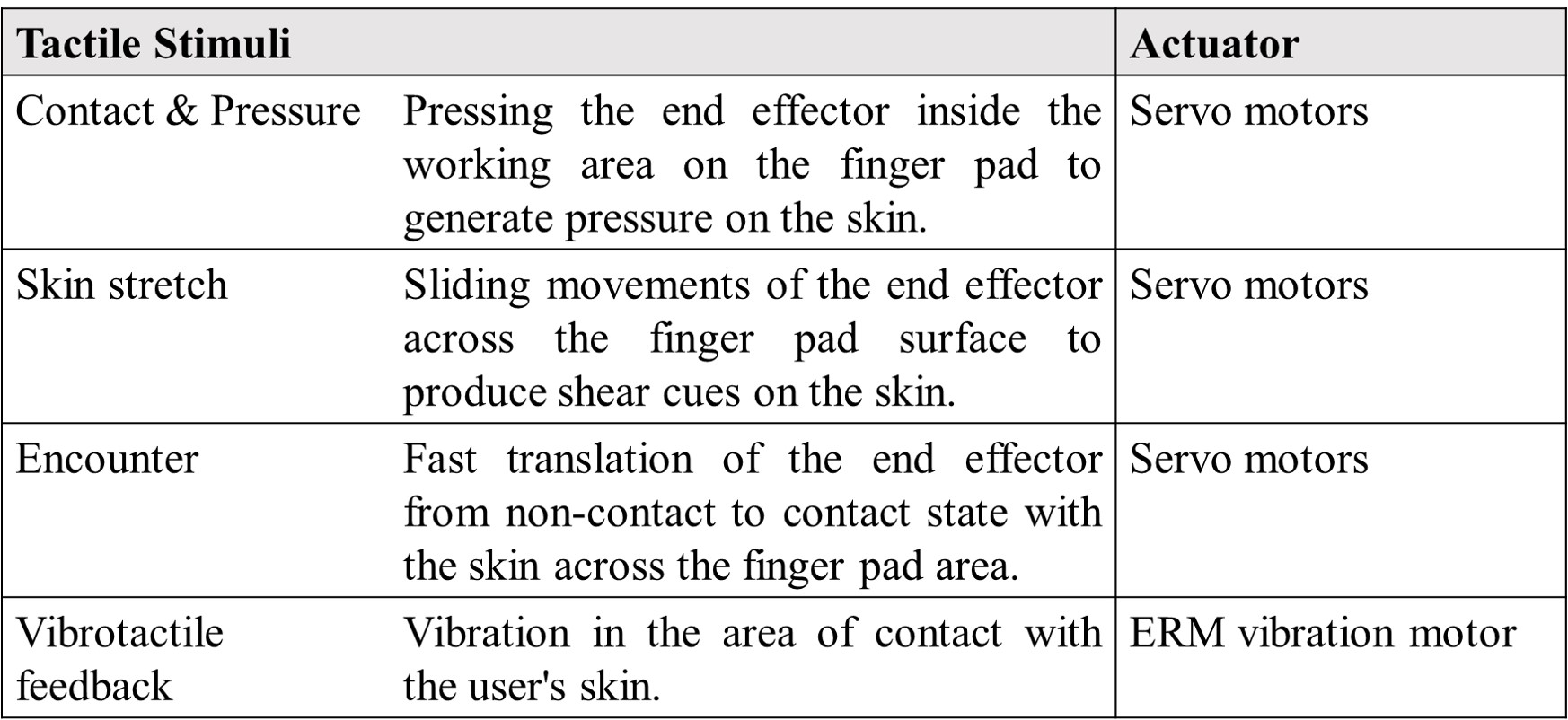}
\end{table} 

FiDTouch display is capable of generating a 3D force vector at the contact point to simulate the normal and shear forces with arbitrary pressure. In addition, the 3-DoF inverted Delta structure allows the end effector to move freely with fast speed inside the workspace while avoiding unnecessary skin contact. Thus, encounter type stimuli can be used to render rapidly changing dynamical interactions with a virtual environment. The vibration motor driven by PWM control is used to generate different vibrotactile cues at the contact point on the finger pad. Thus, FiDTouch could deliver multimodal tactile rendering combining texture simulation (using a vibration motor) and high-speed force feedback.

\section{User Study}
The performance of the developed haptic display was evaluated in a two-stage user study. Firstly, the perception of different contact points simulated on the index finger was explored. Secondly, the ability of users to distinguish the skin stretch directions generated by the end effector of the FiDTouch was tested. In both experiments, the user was asked to sit in front of a desk and to wear the FiDTouch display on the index finger of the dominant hand. During the experiments, the device was hidden from the participants' view and they were provided with visual guides of the tactile patterns presented.

\subsection{Participants}
In total, sixteen subjects (five females) aged from 19 to 30 years (mean=25.2, $SD$=2.7) took part in both experiments, giving their informed consent. The user study procedure was approved by the local Institutional Review Board. Two subjects were left-handed. Before the experiments, the participants had no prior experience using the developed tactile display.

\subsection{User study 1: discrimination of spatial contact points}

\begin{figure}[t]
  \centering
  \includegraphics[width=0.82 \linewidth]{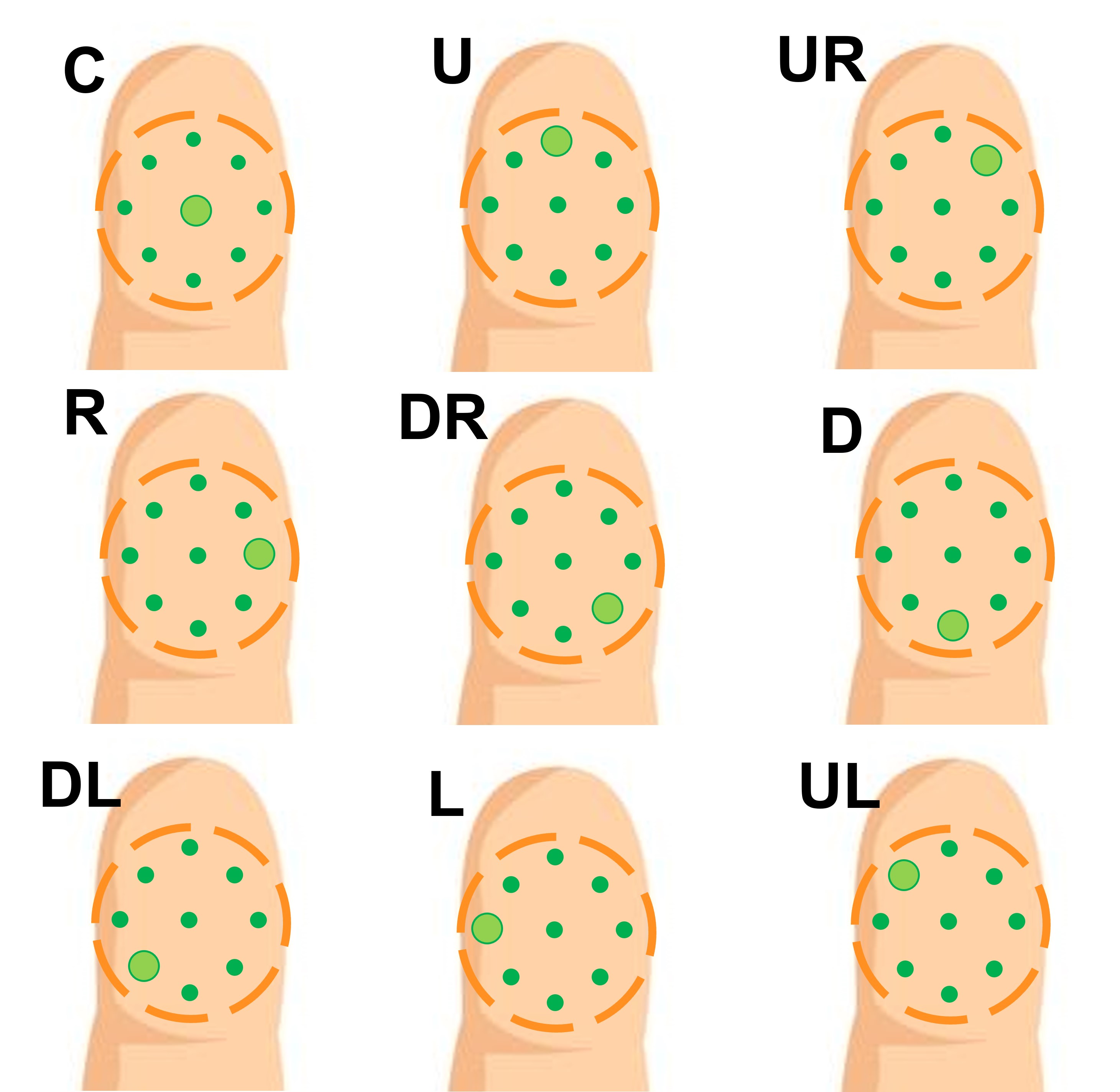}
  \caption{Designed tactile patterns for the discrimination of static contact points. Light-green dots represent contact points, while green dots show the possible contact locations.}
  \label{fig:static patterns}
\end{figure}

The purpose of the experiment was to study the spatial recognition of contact stimulus on the user’s finger pad, differing in the location of the contact point. Nine contact point patterns were designed, distributed across the finger pad within the working area of the FiDTouch display (Fig. \ref{fig:static patterns}). 

\begin{figure} [!t]
\begin{center}
    \subfigure[Distribution of correctly recognized contact points at the index finger pad among all subjects. Average values are marked with crosses. Scatter points represent recognition rate of an individual participant.Statistical significance is marked with an asterisk ($^*: p \leq .05,\ ^{**}:p \leq .01,\ ^{***}:p \leq .001$).]{
   \includegraphics[width=0.9\linewidth]{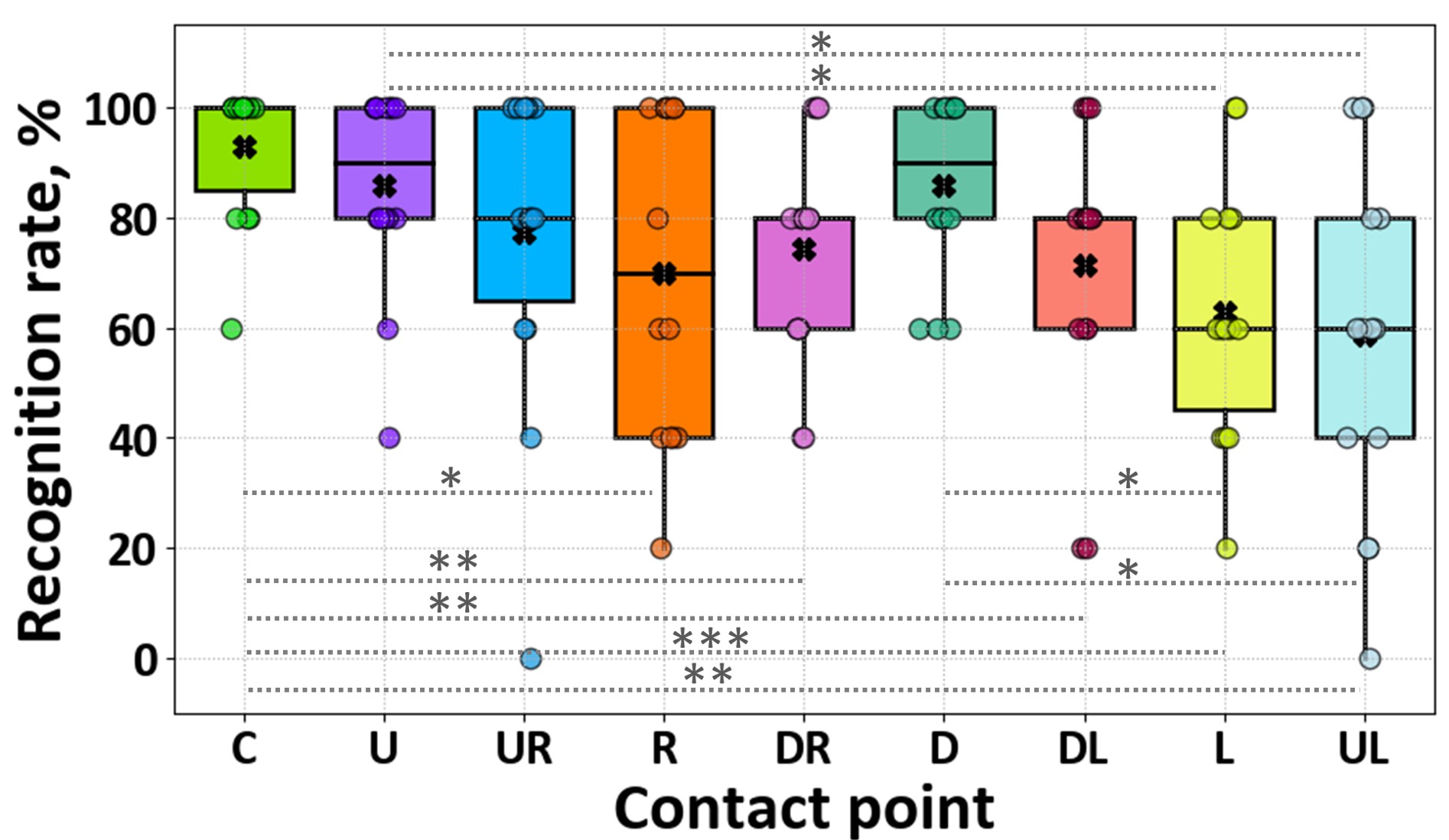}}
    \subfigure[Confusion matrix for actual and perceived contact stimuli across all the subjects.]{
    \includegraphics[width=0.9\linewidth]{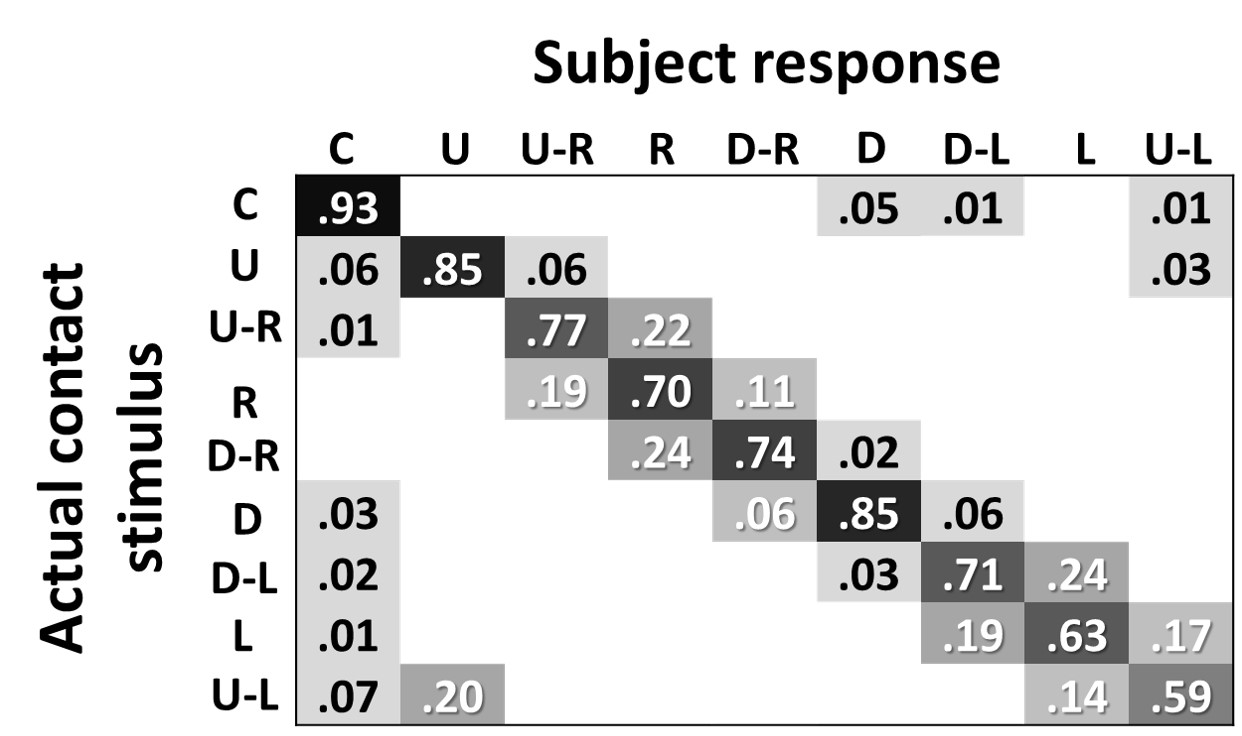}}  
\caption{Experimental results of spatial contact points recognition.}
\label{fig:finger_static_points}
\end{center}
\end{figure} 
\subsubsection{Experimental description}
Before the experiment began, a training session was conducted in which each pattern was presented to the users to test their ability to distinguish the contact state at each location. A calibration of contact force was applied for each contact stimulus to ensure the participant perceived it as comfortable. During the experiment, the participant was provided with a visual guide on the designed patterns. For each trial, the end effector first reached a specified location in a non-contact position (a distance of 5 $mm$ from the finger pad). After that, the end effector moved down to deliver a contact stimulus for half a second, and then returned to the non-contact position. After each trial, the subject was asked to specify the provided contact pattern and the level of his/her confidence. Each pattern was delivered five times in a random order, in total 45 tactile patterns were presented to each subject.

\subsubsection{Experimental results}

The distribution of correctly recognized contact points on the finger pad is shown in Fig. \ref{fig:finger_static_points}(a). The results of the experiment revealed that the average percentage of recognition rates of each contact point averaged across all participants ranged from 59$\%$ to 93$\%$. The mean percentage of correct answers is $75\%$. For all patterns presented, subjects reported an average confidence level of $4.4$, $SD=0.6$. The most recognizable contact positions were \textbf{C}, \textbf{U}, and \textbf{D}, with a recognition rate of $93\%$, $85\%$, and $85\%$, respectively. And the least distinct contact position was \textbf{UL} with a rate of $59\%$. In general, the recognition rate of contact points on the right side of the finger pad was higher than on the left side.

Fig. \ref{fig:finger_static_points}(b) shows a confusion matrix for actual and perceived contact points. For the obtained data, four pairs of patterns (\textbf{UR--R}, \textbf{UL--U}, \textbf{DR--R}, \textbf{DL--L}) for which the recognition error was $20\%$ or more can be identified. It should be noted, that two pairs of contact positions, \textbf{UR--R} and \textbf{DL--L}, have almost symmetrical pairwise confusing. While the recognition errors in \textbf{UL--U}, \textbf{DR--R} pairs were asymmetrical. Thus, the \textbf{UL} contact point was confused with \textbf{U} contact point in $20\%$ of cases; however, the \textbf{U} contact point was confused with \textbf{UL} contact point only in $3\%$ of cases in total.

The experimental results were analyzed using the non-parametric Kruskal-Wallis test, with a chosen significance level of $\alpha<.05$, since the obtained data deviated from the normal distribution.  According to the test findings, there is a statistically significant difference in the recognition rates for the different contact positions $(H = 22.1,p=.005)$. The post-hoc pairwise comparisons were analyzed using the Mann-Whitney U test. It was significantly easier for participants to recognize point \textbf{C} than diagonal contact positions (\textbf{DR}, \textbf{DL}, \textbf{UL}), as well as \textbf{L} and \textbf{R} contact positions. In addition, contact points \textbf{U} and \textbf{D} had significantly higher recognition rates than the points \textbf{L} and \textbf{UL}. Table \ref{statistics_contact} shows the detailed results of the statistical analysis conducted.

\begin{table}[!h]
\caption{Results of Statistical Analysis}
\label{statistics_contact}
\centering
\includegraphics[width=0.98\linewidth]{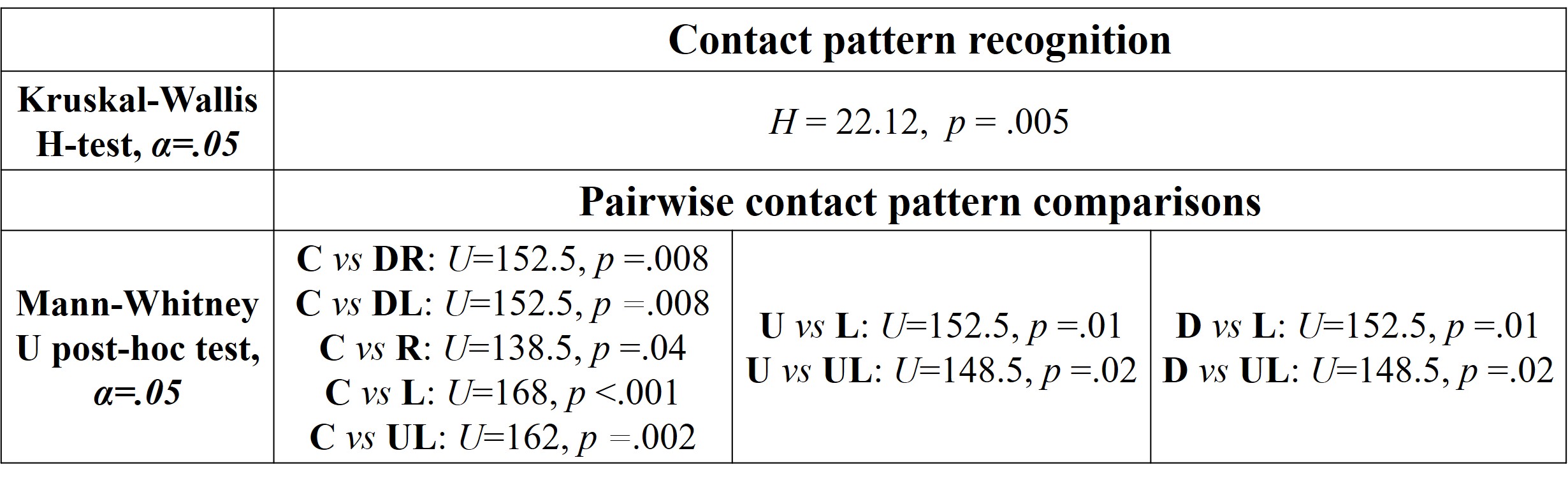}
\end{table}

\subsection{User study 2: discrimination of skin stretch patterns}
\begin{figure}[!h]
  \centering
  \includegraphics[width=0.95 \linewidth]{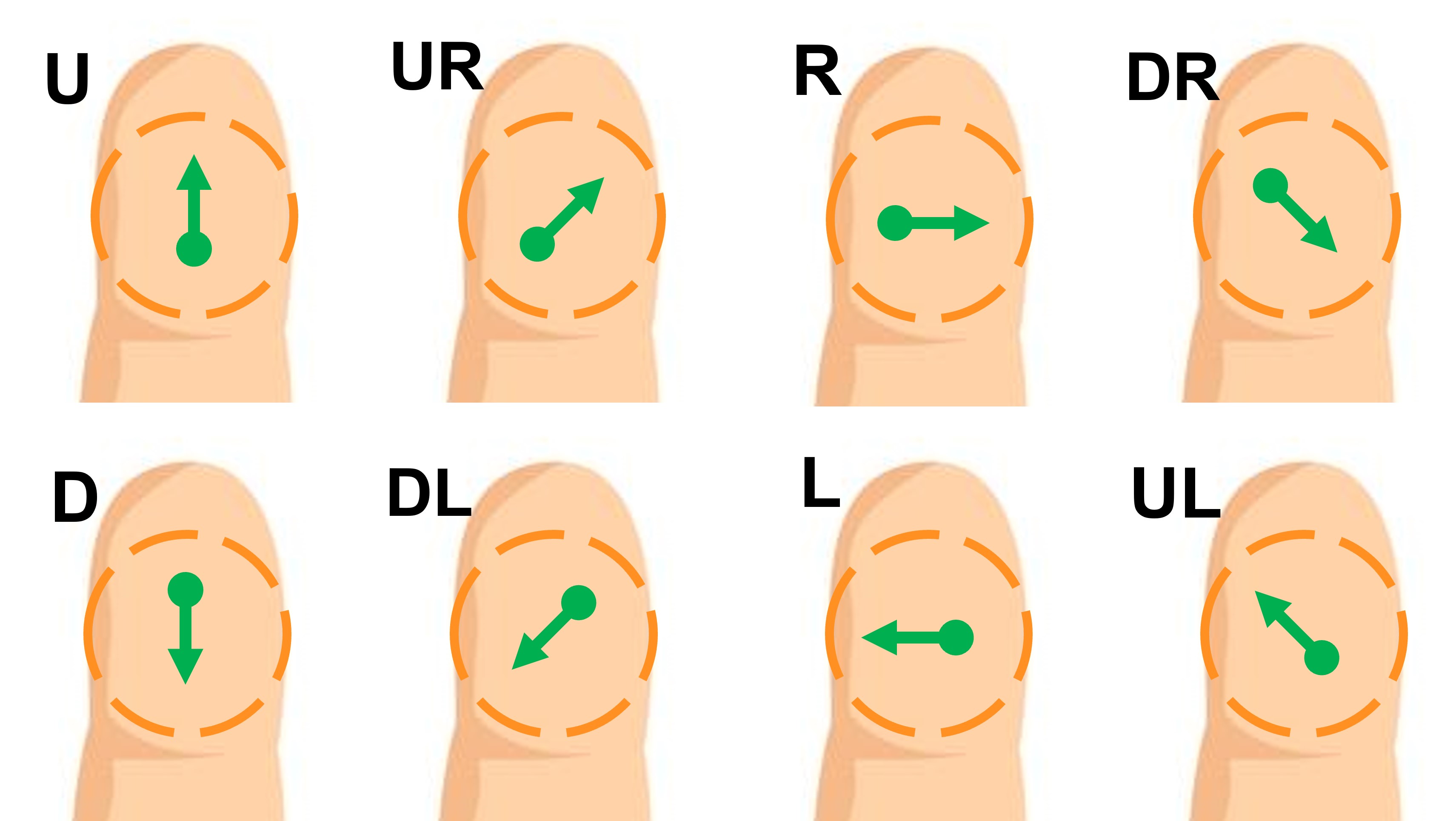}
  \caption{Skin stretch patterns for the experiment. U: upward, D: downward, L: left, R: right, UL: upward to the left, UR: upward to the right, DL: downward to the left, DR: downward to the right. The dot represents the starting contact point, while the arrow means the direction of skin stretch pattern.}
  \label{fig:skin stretch patterns}
\end{figure}

The objective of the experiment was to study the recognition of skin stretch patterns generated on the finger pad. Eight skin stretch patterns were designed, representing the direction of end-effector movement with an increment of 45 $deg.$ (Fig. \ref{fig:skin stretch patterns}).

\subsubsection{Experimental description}
Each pattern was presented five times in random order, therefore, in total 40 patterns were provided to each subject. Before the experiment, each subject had a training session to demonstrate the procedure and to test whether each pattern is well recognized. The design of the device ensures that the area of the finger pad used for stimulation is almost flat, since the most curved area of the fingertip is not involved. For each trial, the end effector translated to a starting location of the skin stretch pattern in a non-contact position (a distance of 5 $mm$ from the finger pad). After that, the end effector moved down to a contact state with a finger pad, translated along a given direction, and then returned to the center non-contact position. After each skin stretch stimulus was delivered, the participant was asked to report the recognized pattern and indicate his or her degree of confidence in the answer on a scale of one to five. 

\subsubsection{Experimental results}
\begin{figure} [!h]
\begin{center}
\includegraphics[width=0.95\linewidth]{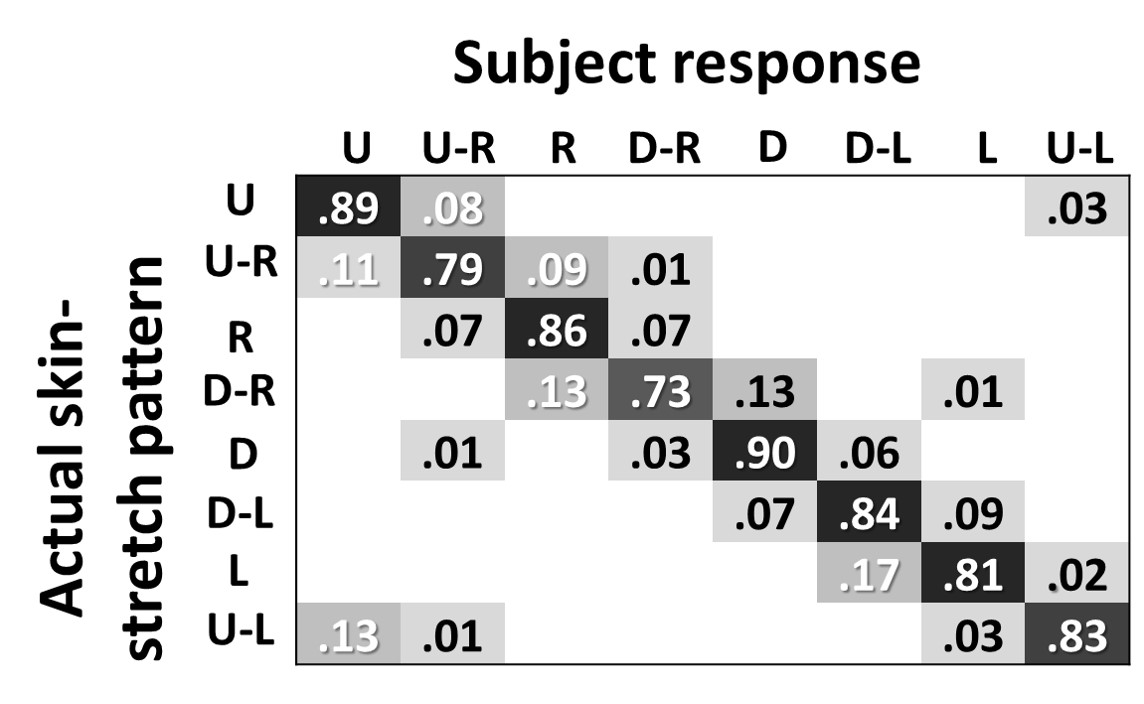}
\caption{Confusion matrix for the actual and perceived skin stretch patterns generated on the finger pad across all subjects.}
\label{fig:skin_stretch_cm}
\end{center}
\end{figure}

The results of the experiment are summarized in a confusion matrix (Fig. \ref{fig:skin_stretch_cm}).

\begin{figure} [!h]
\begin{center}
\includegraphics[width=0.92\linewidth]{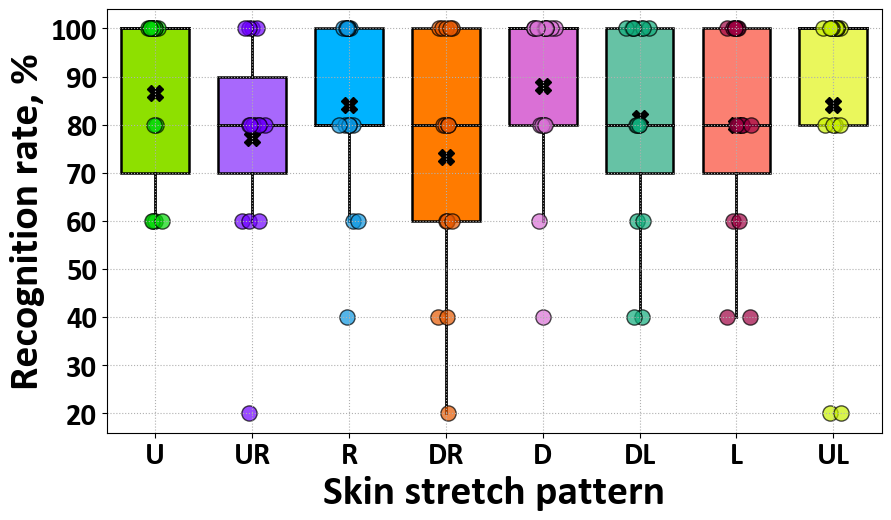}
\caption{Distribution of pattern recognition rate among all subjects. Average values are marked with crosses. Scatter points represent recognition rate of an individual participant.}
\label{fig:skin_stretch}
\end{center}
\end{figure}

In addition, Fig. \ref{fig:skin_stretch} shows the distribution statistics of pattern recognition rates with the representation of individual participant results. Overall, subjects reported an average confidence level of $4.5$ for their responses, $SD = 0.5$. The mean percentage of correct answers comprised 83$\%$. The most distinctive skin stretch patterns were for straight directions such as \textbf{D}, \textbf{U} and \textbf{R} with a recognition rate of $90\%$, $89\%$ and $86\%$, respectively. And the most confusing pattern was \textbf{DR} with a recognition rate of 73$\%$. Thus, \textbf{DR} was confused with patterns \textbf{R} and \textbf{D} in 13$\%$ of the cases. In addition, it can be noted (see Fig. \ref{fig:skin_stretch}) that diagonal skin stretch patterns \textbf{DR}, \textbf{UR} and \textbf{UL} were difficult to recognize clearly for some participants (recognition rate comprised 20\%). The experimental results were analyzed using the non-parametric Kruskal-Wallis test. According to the test findings, there is no statistically significant difference in the recognition rates between skin stretch patterns $(H = 6.3,p>.05)$.

\section{Conclusion}
We have proposed a 3D wearable haptic display to deliver spatial static and dynamic patterns with arbitrary pressure on the surface of the finger pad. The developed finger-worn tactile display was evaluated in a two-stage experiment on the perception of spatial contact stimuli and skin stretch directions generated on the finger pad, which showed high average pattern discrimination rates of 75\% and 83\%, respectively.

The proposed FiDTouch display can augment user interactions in Virtual Environments through multimodal haptic feedback. The haptic display can be applied in medical and rehabilitation systems, teleoperation, and VR simulators. The future work will be aimed at the thorough user study on discrimination of multimodal feedback. We plan to explore the effects of applying various tactile cues to the finger pad areas using the developed display in telemanipulation tasks, such as generating rendering a sense of interaction with rigid and compliance objects, simulating an object slipping out of the gripper to enhance the naturalness of human-robot interaction.

\section{ACKNOWLEDGMENT} 
Research reported in this publication was financially supported by the RSF grant No. 24-41-02039.

\bibliographystyle{IEEEtran}
\balance
\bibliography{ref}

\begin{thebibliography}{10}
\providecommand{\url}[1]{#1}
\csname url@samestyle\endcsname
\providecommand{\newblock}{\relax}
\providecommand{\bibinfo}[2]{#2}
\providecommand{\BIBentrySTDinterwordspacing}{\spaceskip=0pt\relax}
\providecommand{\BIBentryALTinterwordstretchfactor}{4}
\providecommand{\BIBentryALTinterwordspacing}{\spaceskip=\fontdimen2\font plus
\BIBentryALTinterwordstretchfactor\fontdimen3\font minus \fontdimen4\font\relax}
\providecommand{\BIBforeignlanguage}[2]{{%
\expandafter\ifx\csname l@#1\endcsname\relax
\typeout{** WARNING: IEEEtran.bst: No hyphenation pattern has been}%
\typeout{** loaded for the language `#1'. Using the pattern for}%
\typeout{** the default language instead.}%
\else
\language=\csname l@#1\endcsname
\fi
#2}}
\providecommand{\BIBdecl}{\relax}
\BIBdecl

\bibitem{gil2022handguidance}
G.~D. Gil, N.~Zemiti, J.~M. Walker, F.~P{\'e}chereau, and P.~Poignet, ``Enabling 4-dof hand guidance using a portable haptic device exerting tangential force on the user's finger pads,'' \emph{Mechatronics}, vol.~86, p. 102868, 2022.

\bibitem{zhu2022cutaneous}
Y.~Zhu, J.~Colan, T.~Aoyama, and Y.~Hasegawa, ``Cutaneous feedback interface for teleoperated in-hand manipulation,'' in \emph{2022 IEEE/RSJ Int. Conf. on Intelligent Robots and Systems (IROS)}, 2022, pp. 605--611.

\bibitem{tsykunov2019swarmtouch}
E.~Tsykunov, R.~Agishev, R.~Ibrahimov, L.~Labazanova, A.~Tleugazy, and D.~Tsetserukou, ``Swarmtouch: Guiding a swarm of micro-quadrotors with impedance control using a wearable tactile interface,'' \emph{IEEE transactions on haptics}, vol.~12, no.~3, pp. 363--374, 2019.

\bibitem{leonardis20173}
D.~Leonardis, M.~Solazzi, I.~Bortone, and A.~Frisoli, ``A 3-rsr haptic wearable device for rendering fingertip contact forces,'' \emph{IEEE Transactions on Haptics}, vol.~10, no.~3, pp. 305--316, 2017.

\bibitem{schorr2017three}
S.~B. Schorr and A.~M. Okamura, ``Three-dimensional skin deformation as force substitution: Wearable device design and performance during haptic exploration of virtual environments,'' \emph{IEEE Transactions on Haptics}, vol.~10, no.~3, pp. 418--430, 2017.

\bibitem{lee2018wearable}
Y.~Lee, M.~Kim, Y.~Lee, J.~Kwon, Y.-L. Park, and D.~Lee, ``Wearable finger tracking and cutaneous haptic interface with soft sensors for multi-fingered virtual manipulation,'' \emph{IEEE/Asme Transactions on Mechatronics}, vol.~24, no.~1, pp. 67--77, 2018.

\bibitem{young2019implementation}
E.~M. Young and K.~J. Kuchenbecker, ``Implementation of a 6-{DOF} {P}arallel {C}ontinuum {M}anipulator for {D}elivering {F}ingertip {T}actile {C}ues,'' \emph{IEEE transactions on haptics}, vol.~12, no.~3, pp. 295--306, 2019.

\bibitem{tzemanaki2018design}
A.~Tzemanaki, G.~A. Al, C.~Melhuish, and S.~Dogramadzi, ``Design of a wearable fingertip haptic device for remote palpation: Characterisation and interface with a virtual environment,'' \emph{Frontiers in Robotics and AI}, vol.~5, p.~62, 2018.

\bibitem{chinello2019modular}
F.~Chinello, M.~Malvezzi, D.~Prattichizzo, and C.~Pacchierotti, ``A {M}odular {W}earable {F}inger {I}nterface for {C}utaneous and {K}inesthetic {I}nteraction: {C}ontrol and {E}valuation,'' \emph{IEEE Transactions on Industrial Electronics}, vol.~67, no.~1, pp. 706--716, 2019.

\bibitem{linkring}
A.~Ivanov, D.~Trinitatova, and D.~Tsetserukou, ``Linkring: a wearable haptic display for delivering multi-contact and multi-modal stimuli at the finger pads,'' in \emph{in Proc. Int. Conf. EuroHaptics}, 2020, pp. 434--441.

\bibitem{williams20214}
S.~R. Williams, J.~M. Suchoski, Z.~Chua, and A.~M. Okamura, ``A 4-dof parallel origami haptic device for normal, shear, and torsion feedback,'' \emph{arXiv preprint arXiv:2109.12134}, 2021.

\bibitem{haptigami}
F.~H. Giraud, S.~Joshi, and J.~Paik, ``Haptigami: A fingertip haptic interface with vibrotactile and 3-dof cutaneous force feedback,'' \emph{IEEE Transactions on Haptics}, vol.~15, no.~1, pp. 131--141, 2021.

\bibitem{fingerprint}
Z.~Zhakypov and A.~M. Okamura, ``Fingerprint: A 3-d printed soft monolithic 4-degree-of-freedom fingertip haptic device with embedded actuation,'' in \emph{in Proc. IEEE Int. Conf. on Soft Robotics (RoboSoft)}, 2022, pp. 938--944.

\bibitem{hummel2016lightweight}
J.~Hummel, J.~Dodiya, G.~A. Center, L.~Eckardt, R.~Wolff, A.~Gerndt, and T.~W. Kuhlen, ``A {L}ightweight {E}lectrotactile {F}eedback {D}evice for {G}rasp {I}mprovement in {I}mmersive {V}irtual {E}nvironments,'' in \emph{in Proc. IEEE Int. Conf. on Virtual Reality and 3D User Interfaces (VR)}, 2016, pp. 39--48.

\bibitem{yem2018effect}
V.~Yem, K.~Vu, Y.~Kon, and H.~Kajimoto, ``Effect of {E}lectrical {S}timulation {H}aptic {F}eedback on {P}erceptions of {S}oftness-{H}ardness and {S}tickiness {W}hile {T}ouching a {V}irtual {O}bject,'' in \emph{in Proc. IEEE Int. Conf. on Virtual Reality and 3D User Interfaces (VR)}, 2018, pp. 89--96.

\bibitem{withana2018tacttoo}
A.~Withana, D.~Groeger, and J.~Steimle, ``Tacttoo: A thin and feel-through tattoo for on-skin tactile output,'' in \emph{in Proc. of ACM Symposium on User Interface Software and Technology (UIST)}, 2018, pp. 365--378.

\bibitem{vizcay2021electrotactile}
S.~Vizcay, P.~Kourtesis, F.~Argelaguet, C.~Pacchierotti, and M.~Marchal, ``Electrotactile feedback for enhancing contact information in virtual reality,'' in \emph{Int. Conf. on Artificial Reality and Telexistence and Eurographics Symposium on Virtual Environments (ICAT-EGVE )}, 2021.

\bibitem{miyakami2019hapballoon}
M.~Miyakami, K.~A. Murata, and H.~Kajimoto, ``Hapballoon: Wearable haptic balloon-based feedback device,'' in \emph{SIGGRAPH Asia 2019 Emerging Technologies}, 2019, pp. 17--18.

\bibitem{han2018hydroring}
T.~Han, F.~Anderson, P.~Irani, and T.~Grossman, ``Hydroring: Supporting mixed reality haptics using liquid flow,'' in \emph{Proc. of ACM Symposium on User Interface Software and Technology (UIST)}, 2018, pp. 913--925.

\bibitem{talhan2022multi}
A.~Talhan, S.~Kumar, H.~Kim, W.~Hassan, and S.~Jeon, ``Multi-mode soft haptic thimble for haptic augmented reality based application of texture overlaying,'' \emph{Displays}, vol.~74, p. 102272, 2022.

\bibitem{ma2024airpush}
Y.~Ma, T.~Xie, P.~Zhang, H.~Kim, and S.~Je, ``Airpush: A pneumatic wearable haptic device providing multi-dimensional force feedback on a fingertip,'' in \emph{in Proc. ACM Int. Conf. on Human Factors in Computing Systems (CHI)}, 2024, pp. 1--13.

\bibitem{hring}
C.~Pacchierotti, G.~Salvietti, I.~Hussain, L.~Meli, and D.~Prattichizzo, ``The hring: A wearable haptic device to avoid occlusions in hand tracking,'' in \emph{in Proc. IEEE Haptics Symposium (HAPTICS)}, 2016, pp. 134--139.

\bibitem{de2018enhancing}
X.~De~Tinguy, C.~Pacchierotti, M.~Marchal, and A.~Lecuyer, ``Enhancing the stiffness perception of tangible objects in mixed reality using wearable haptics,'' in \emph{in Proc. IEEE Int. Conf. on Virtual Reality and 3D User Interfaces (VR)}, 2018.

\bibitem{preechayasomboon2021haplets}
P.~Preechayasomboon and E.~Rombokas, ``Haplets: Finger-worn wireless and low-encumbrance vibrotactile haptic feedback for virtual and augmented reality,'' \emph{Frontiers in Virtual Reality}, vol.~2, p. 738613, 2021.

\bibitem{kawazoe2021tactile}
A.~Kawazoe, G.~Reardon, E.~Woo, M.~Di~Luca, and Y.~Visell, ``Tactile echoes: Multisensory augmented reality for the hand,'' \emph{IEEE Transactions on Haptics}, vol.~14, no.~4, pp. 835--848, 2021.

\bibitem{maeda2022fingeret}
T.~Maeda, S.~Yoshida, T.~Murakami, K.~Matsuda, T.~Tanikawa, and H.~Sakai, ``Fingeret: A wearable fingerpad-free haptic device for mixed reality,'' in \emph{Symposium on Spatial User Interaction}, 2022, pp. 1--10.

\bibitem{moriyama2022wearable}
T.~Moriyama and H.~Kajimoto, ``Wearable {H}aptic {D}evice {P}resenting {S}ensations of {F}ingertips to the {F}orearm,'' \emph{IEEE Transactions on Haptics}, vol.~15, no.~1, pp. 91--96, 2022.

\bibitem{palmer2022haptic}
J.~E. Palmer, M.~Sarac, A.~A. Garza, and A.~M. Okamura, ``Haptic feedback relocation from the fingertips to the wrist for two-finger manipulation in virtual reality,'' in \emph{2022 IEEE/RSJ Int. Conf. on Intelligent Robots and Systems (IROS)}, 2022, pp. 628--633.

\bibitem{tanaka2024reawristic}
Y.~Tanaka, N.~Weiss, R.~C. Bolger-Cruz, J.~Hartcher-O’Brien, B.~Flynn, R.~Boldu, and N.~Colonnese, ``Reawristic: Remote touch sensation to fingers from a wristband via visually augmented electro-tactile feedback,'' in \emph{2024 IEEE Int. Symp. on Mixed and Augmented Reality (ISMAR)}, 2024, pp. 951--960.

\bibitem{deltatouch}
D.~Trinitatova and D.~Tsetserukou, ``Deltatouch: a 3d haptic display for delivering multimodal tactile stimuli at the palm,'' in \emph{in Proc. IEEE Int. Conf. World Haptics (WHC)}, 2019, pp. 73--78.

\bibitem{touchvrr}
------, ``Touchvr: A wearable haptic interface for vr aimed at delivering multi-modal stimuli at the user’s palm,'' in \emph{SIGGRAPH Asia 2019 XR}, 2019, pp. 42--43.

\end{thebibliography}
\end{document}